\documentclass[a4paper,french]{rnti}

\usepackage[T1]{fontenc}
\usepackage[utf8]{inputenc}
\usepackage{amssymb} 
\usepackage{multirow} 
\usepackage{amsmath} 
\usepackage{subfig}
\usepackage{xcolor} 
\usepackage{url}

\usepackage{graphicx}

\titrecourt{Étude comparative de modèles de clustering de séries temporelles}

\nomcourt{V. Courrier et al.}

\titre{Étude comparative de modèles de clustering de séries temporelles multivariées issues d'objets médicaux connectés}

\auteur{Violaine Courrier\affil{1},
        Christophe Biernacki\affil{2}, 
        Cristian Preda\affil{2},
        Benjamin Vittrant\affil{3}
        }

\affiliation{
    \affil{1} Inria, Univ. Lille / Withings,
          violaine.courrier@inria.fr\\
    
    \affil{2}Inria, Univ. Lille,
          \{christophe.biernacki, cristian.preda\}@inria.fr \\
    
    \affil{3} Withings,
          benjamin.vittrant@withings.com\\
 }

\resume{%
Dans le domaine de la santé, les données des patients sont souvent collectées sous forme de séries temporelles multivariées, offrant une vue complète de l'état de santé d'un patient au fil du temps. Ces données sont généralement éparses et épisodiques. Cependant, les objets médicaux connectés peuvent augmenter la fréquence des données. L'objectif est de créer de manière non supervisée des profils de patients à partir de ces séries temporelles. En l'absence de labels, un modèle prédictif peut être utilisé pour prédire les valeurs futures tout en formant un espace de clusters latents, évalué en fonction de la performance prédictive. 
À l'aide des données réelles de l'entreprise Withings, nous comparons les approches de clustering statique \textsc{MagmaClust}, qui crée un cluster à l'échelle de toute la série temporelle, et de clustering dynamique \textsc{DGM²}, qui permet à l'appartenance d'un individu à un groupe de changer avec le temps.
}

\summary{%
In healthcare, patient data is often collected as multivariate time series, providing a comprehensive view of a patient's health status over time. While this data can be sparse, connected devices may enhance its frequency. The goal is to create patient profiles from these time series. In the absence of labels, a predictive model can be used to predict future values while forming a latent cluster space, evaluated based on predictive performance. We compare two models on Withing's datasets, \textsc{MagmaClust} which clusters entire time series and \textsc{DGM²} which allows the group affiliation of an individual to change over time (dynamic clustering).
}

\begin{document}

\section{Introduction}

Dans le domaine de la santé, les données des patients sont souvent recueillies sous forme de séries temporelles multivariées (MTS, pour Multidimensional Time Series). Cela signifie que plusieurs paramètres sont mesurés à différents moments pour chaque patient (sa tension artérielle, sa fréquence cardiaque, son poids, etc.). Ces données sont extrêmement précieuses car elles permettent de suivre l'évolution de l'état de santé d'un patient dans le temps et de détecter des indicateurs d’une maladie ou d’un risque de maladie. Une contrainte majeure réside dans la sparsité de ces données. En effet, le patient n’est que rarement à l’hôpital (pour des examens par exemple), ce qui signifie que les mesures enregistrées relatives à sa santé sont essentiellement épisodiques. Cependant, des objets médicaux connectés offrent un suivi à domicile du patient, ce qui permet d’augmenter la fréquence des mesures. Ces mesures à domicile ne sont en revanche pas toujours réalisées dans des conditions contrôlées, ce qui peut affecter leur fiabilité. \\

Le grand apport des séries temporelles multivariées, en plus de la dimension temporelle, est le caractère multivarié qui offre une vue plus complète et plus précise de l'état de santé d'un patient. En analysant plusieurs variables simultanément, cela permet d'utiliser les effets de corrélation et de causalité entre différentes variables, des informations qui seraient perdues dans une analyse univariée. Cela peut aider à identifier des facteurs de risque ou des signes précoces de maladie qui pourraient être manqués autrement.

\section{Clustering de séries temporelles multivariées}
L’objectif est de créer des profils patients à partir de séries temporelles multivariées qui illustrent leur état de santé. Une particularité de ces données est l’absence de labels. Par exemple, un patient atteint d'hypertension ne le précise pas au moment de sa prise de tension ou bien lors de sa pesée. Son état de santé est a priori inconnu, et il s'agit de le déterminer. Cela revient alors à créer des groupes de patients de façon non supervisée. Un intérêt de ces clusters est de faire de la prévention, en ciblant par exemple les individus d'un cluster où l'incidence d'une maladie est significativement plus élevée. 

\subsection{Etat de l'art}

Le clustering de séries temporelles est un sujet majeur. Une revue de la littérature par \cite{ts_clustering_survey} identifie 3 catégories de méthodes : le "clustering de séries temporelles entières", le "clustering de sous-séquences" et le "clustering de points temporels", que l'on appellera dans cet article le "clustering dynamique". Selon \citet{preda_survey}, il y a quatre familles de modèles de clustering au niveau de la série temporelle entière. \\

La première famille traite les séries temporelles comme des entités de haute dimension, en négligeant leur nature temporelle, une approche qui peut être illustrée par les travaux de \citet{hdclustering}. La seconde stratégie adopte une démarche en deux temps : une réduction de dimension initiale, par des techniques telles que l'Analyse en Composantes Principales (ACP), suivie d'un clustering classique. La troisième famille repose sur des approches non paramétriques, qui privilégient l'utilisation de mesures de distance ou de dissimilarité spécifiques aux séries temporelles, telles que la distance de Dynamic Time Warping (DTW), pour ensuite appliquer des algorithmes de clustering traditionnels adaptés aux données de dimension finie (exemple donné par \cite{ex_kmeans_dtw}). La dernière catégorie englobe les méthodes qui supposent une distribution de probabilité sous-jacente aux données, souvent mises en œuvre via des modèles génératifs ou des techniques de clustering basées sur des mélanges de distributions. \\

Les développements récents dans ce domaine incluent l'intégration du deep learning (par exemple l'article de \cite{TSclusterinngdeep}), qui ouvrent de nouvelles perspectives pour le clustering de séries temporelles de grande dimension.

\subsection{Comparaison DGM2 et MagmaClust}
Dans le contexte où les données ne sont pas labellisées, nous proposons d'utiliser un modèle prédictif qui génère un espace de clustering latent. Ce modèle positionne les individus ayant des caractéristiques similaires dans le même groupe et exploite ces informations pour améliorer la précision de la prédiction. L'utilisation d'un modèle prédictif offre l'avantage d'évaluer la qualité du clustering en se basant sur la performance prédictive du modèle, mesurée par des indicateurs tels que l'erreur quadratique moyenne (RMSE) ou l'erreur absolue moyenne (MAE). Nous partons du principe qu'un clustering plus précis et pertinent fournit des informations supplémentaires qui améliorent la capacité de prédiction du modèle. \\

Nous présentons deux méthodes de clustering distinctes : la première réalise un clustering de la série temporelle dans son intégralité, tandis que la seconde adopte une approche de clustering dynamique, attribuant un cluster unique à chaque instant temporel. Ces deux méthodes sont basées sur des modèles gaussiens : l'un utilise des processus gaussiens dans une modélisation bayesienne tandis que l'autre repose sur une distribution de mélange gaussien. L'adoption d'une base gaussienne pour ces modèles nous intéresse pour sa flexibilité et sa capacité à capturer des structures sous-jacentes des données. \\

Dans la suite, on notera $X_i$ la série temporelle multivariée d’un individu $i$, avec $X_i = X_{i,1:T}=(X_{i,1},…,X_{i,T})$, où $X_{i,t} = (X^1_{i,t}, ..., X^d_{i,t})$ avec $X^j_{i,t}$ la valeur de la variable $j$ de l'individu $i$ au temps $t$, $\forall j=1,...,d, \ t = 1,...,T$. 

\subsection{Clustering statique de séries entières}

\begin{figure}[t]
\begin{center}
 \includegraphics[width=10cm]{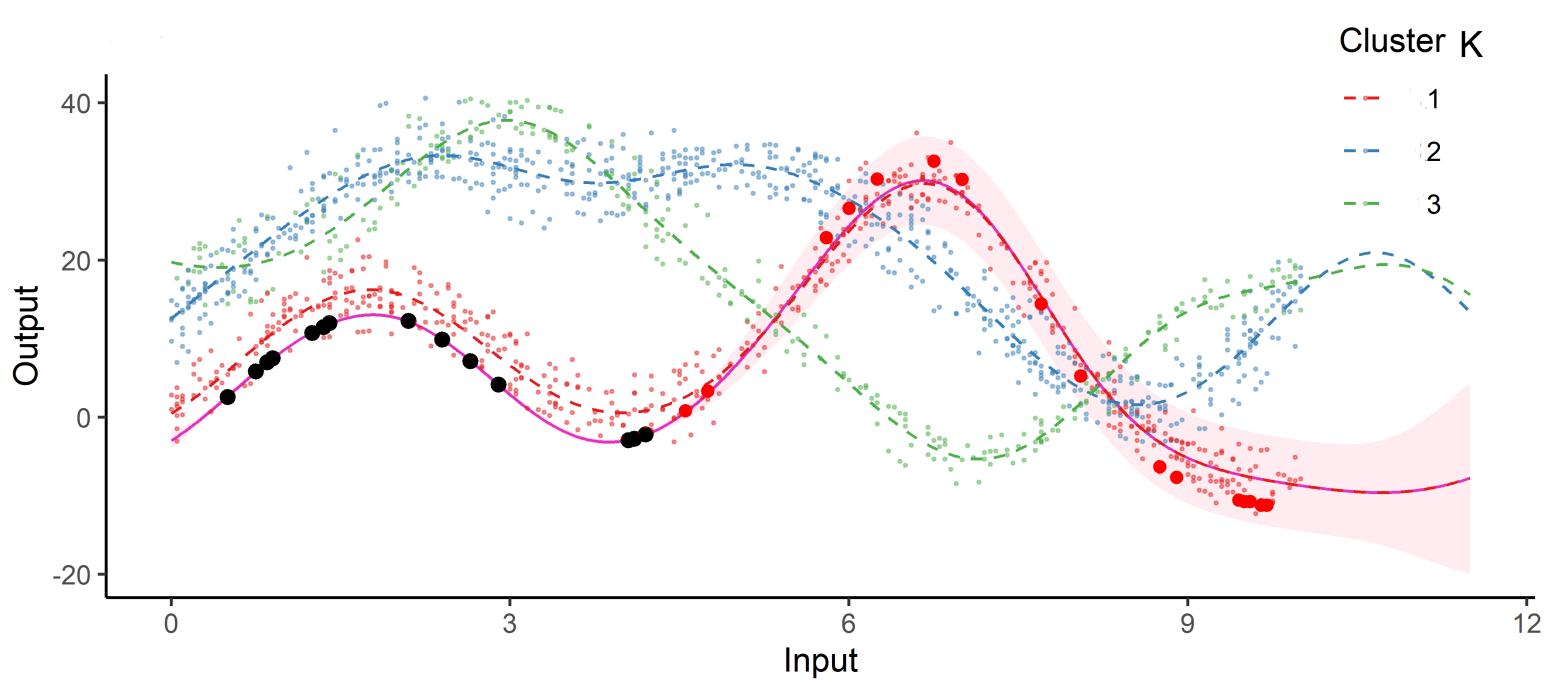}
 \caption{Courbes de prédiction (violet) avec les intervalles de crédibilité à 95 \% associés (rose) de \textsc{MagmaClust}. Les lignes en pointillé représentent les moyennes des estimations des processus moyens. Les points de données observés sont en noir, les points de données de test sont en rouge. Les points en arrière plans sont les observations de l'ensemble des données d'apprentissage, colorées par rapport à leur appartenance à un cluster. "Input" correspond aux pas de temps et "Output" à la valeur de la variable.} \label{magmaclust}
\end{center}
\end{figure}

En partant de l’hypothèse d'existence d'une structure de groupe dans les données, nous proposons d'utiliser l'extension d'un modèle de prédiction (\textsc{Magma}, \citet{magma}). Ce modèle introduit une étape de clustering à $K$ groupes à l’aide d’un mélange de processus gaussiens (GP, pour Gaussian Process) (\textsc{MagmaClust} de \citet{magmaclust}) \footnote{https://github.com/ArthurLeroy/MagmaClustR}. Ce modèle utilise plusieurs processus gaussiens pour modéliser la moyenne des données. Le modèle génératif proposé se définit alors pour le cluster $k$, $k=1,...,K$, comme suit :
$$
X_{i,t} = \mu_{k,t} + f_{i,t} + \epsilon_{i,t}, \forall i, \forall t \in \{1,...,T\}
$$
avec 
\begin{itemize}
    \item $\mu_{k,t}$ le GP moyen spécifique au $k$-ème groupe,
    \item $f_{i,t}$ le GP spécifique à l’individu $i$,
    \item $\epsilon_{i,t} $ le GP du bruit spécifique à l’individu $i$.
\end{itemize} 
\ \\
Cette modélisation correpond à un modèle mixte avec la moyenne du groupe comme effet fixe et l'effet sujet et l'erreur en effets aléatoires. Des hypothèses d'indépendance entre ces processus sont considérées. \\

Pour illustrer le modèle, la Figure \ref{magmaclust} montre une prédiction pour le cluster le plus probable (ici le cluster 1 parmis $K = 3$ classes) à partir de données synthétiques.

\subsection{Clustering dynamique}

Au vu de l’historique d'enregistrement des mesures qui peut être long, et puisque l’état de santé du patient évolue dans le temps, il est souhaitable que l’appartenance d’un individu à un groupe puisse également évoluer dans le temps. Le clustering par intervalle de temps serait un candidat (modèles ‘interval-based’) mais il nécessite de définir le début et la fin de la séquence d’intérêt. On s'intéresserait plutôt dans notre cas à classifier par pas de temps, et donc à faire ce qu’on appellera ici du clustering dynamique, c’est-à-dire que l’appartenance d’un individu à un cluster est indexée par le temps. \\

\begin{figure}[t]
\begin{center}
 \includegraphics[width=9cm]{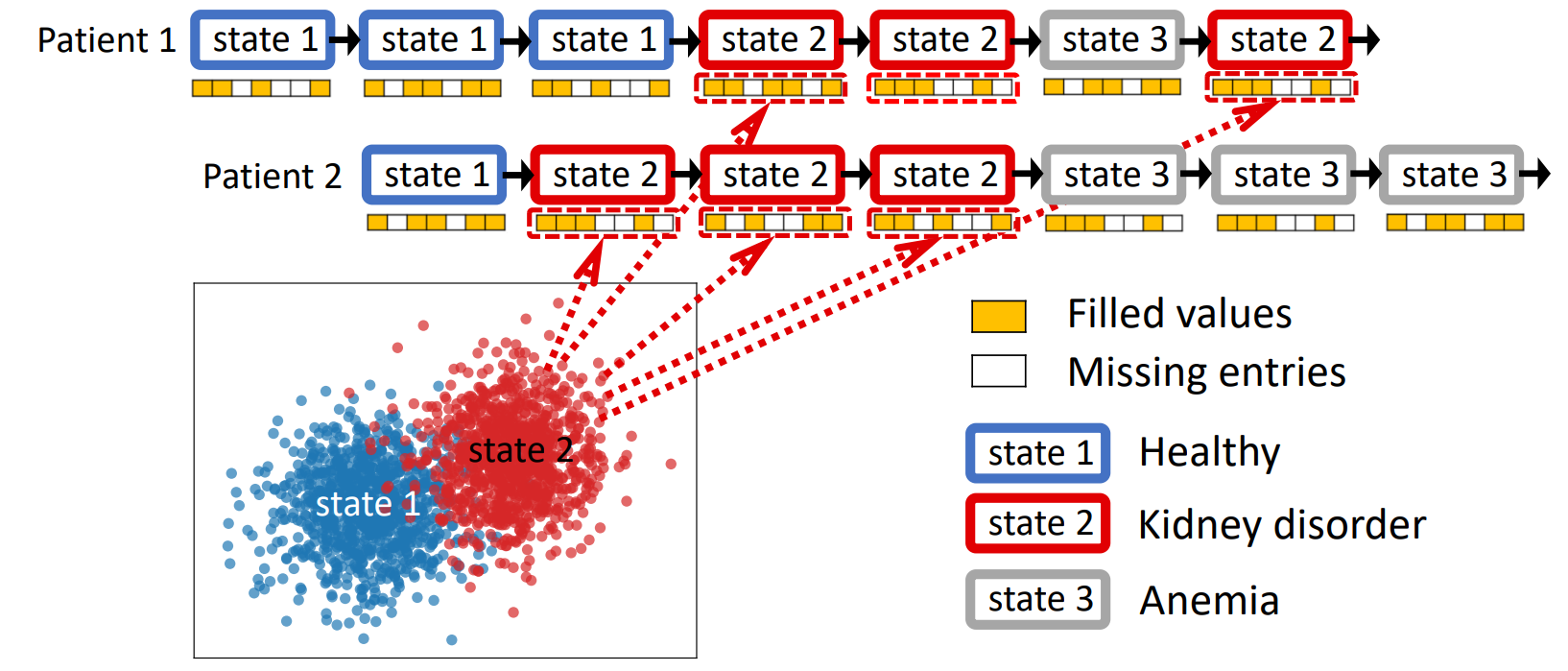}
 \caption{Illustration de clustering dynamique des MTS éparses de deux patients dialysés. Le vecteur sous chaque état est une caractéristique temporelle générée à partir d'une certaine distribution.} \label{dyalisis}
\end{center}
\end{figure}

Par exemple, considérons les MTS collectés pour les patients en dialyse. La dialyse est une thérapie de remplacement rénal importante pour purifier le sang des patients dont les reins ne fonctionnent pas normalement. Les patients en dialyse ont des routines qui entraînent la collecte d'un nombre varié de signaux. Un modèle qui associe un état de santé à chaque pas de temps, comme illustré dans la Figure \ref{dyalisis}, pourrait aider les professionnels de la santé à détecter rapidement les changements significatifs dans l'état du patient. \\

Le clustering dynamique suppose l'exsitence d'un processus qualitatif latent à $K$ modalités, $Z =\{Z_t, t \in [0,T]\}$, avec $Z_{t}\in \{1,\ldots, K\}$. Ainsi, à chaque série $X_{i,t}$, on associe la trajectoire qualitative définissant le groupe du patient $i$, $z_{i,t} \in \{1,...K\}$.  \\

Le modèle \textsc{DGM²} de \citet{DGM2} \footnote{https://github.com/thuwuyinjun/DGM2} permet de faire ce clustering dynamique. C’est un modèle génératif qui suit la transition des clusters latents afin d'obtenir une modélisation robuste. Ce modèle se distingue par une distribution de mélange gaussien dynamique, qui saisit la dynamique des structures de clustering. Les auteurs introduisent un modèle génératif qui capture les structures latentes de regroupement dynamique pour des prévisions robustes. Il suit le cadre de transition et d'émission suivant :
\begin{enumerate}
    \item Probabilité de transition : la probabilité d'un nouvel état $z_{i, t+1}$ est mise à jour en fonction des états précédents $z_{i, 1:t} = (z_{i,1},…,z_{i,t})$, c'est-à-dire 
    \begin{align*}
        p(z_{i, t+1}|z_{i, 1:t}) & = softmax(MLP(h_{i,t})), \\
        \text{avec } h_{i,t} & = RNN(z_{i, t}, h_{i,t-1}).
    \end{align*}
    MLP représente un perceptron multicouche (Multilayer Perceptron) et RNN des réseaux neuronaux récurrents (Recurrent Neural Networks).
    \item Tirer $z_{i,t+1}$ d'une distribution catégorielle sur toutes les composantes du mélange c'est-à-dire $z_{i,t+1} \sim \mathcal{M}(p(z_{i, t+1}|z_{i, 1:t}))$.
    \item Tirer $\tilde{z}_{i,t+1}$ du mélange de distribution gaussiennes dynamique c'est-à-dire $\tilde{z}_{i,t+1} \sim \mathcal{M}(\psi_{i,t+1})$ (voir l'annexe \ref{dynamique} pour le détail de $\psi_{i,t+1}$).
    \item Tirer $\tilde{x}_{i,t+1}$ à partir de la distribution gaussienne du cluster auquel appartient $\tilde{z}_{i,t+1}$.
\end{enumerate}
$z_{i,t+1}$ est est utilisé dans la transition (étape 1) pour maintenir la propriété récurrente ; $\tilde{z}_{i,t+1}$ est utilisé dans l'émission (étape 4). \\

Pour approcher la distribution a posteriori, paramétrée par des réseaux de neurones avec des paramètres $\phi$, on utilise le modèle d'inférence suivant :
\begin{align*}
    q_\phi(z_{t+1}|x_{1:t+1}, z_{t}) &= softmax(MLP(\tilde{h}_{t+1})), \\
    \text{avec } \tilde{h}_{t+1} &= RNN(x_t, \tilde{h}_t).
\end{align*}

Dans notre expérience, les deux RNNs sont des LSTMs (Long Short-Term Memory, introduit par \cite{lstm}).

\section{Comparaison des méthodes sur des données réelles}

Nous menons une analyse comparative des performances de \textsc{MagmaClust} et \textsc{DGM²}, et nous invitons les lecteurs à consulter les publications dédiées à ces modèles pour une comparaison de leurs performances avec les méthodes de l'état de l'art. 

\subsection{Protocole experimental}
\paragraph{Présentation des données} Pour comparer les deux modèles précédents, nous considèrons les données de l’entreprise Withings \footnote{https://www.withings.com/fr/fr/}. Withings est une entreprise qui conçoit des objets médicaux connectés tels que des montres, des balances ou encore des tensiomètres. 
\\

Nous avons créé un dataset de données agrégées par mois pour l'année 2022. Les utilisateurs ont de 18 à 100 ans, et possédent une balance et un sleep analyzer\footnote{Un appareil placé sous le matelas qui permet de récupérer des informations sur la nuit de l'utilisateur.}. Nous nous intéressons à 2 variables : \textit{BMI} (la moyenne par mois du BMI\footnote{BMI = poids[kg] / taille²[m²]}), et \textit{Sleep} (la moyenne par mois de la durée de sommeil). Nous avons établi des critères selon lesquels une nuit de sommeil doit durer entre 45 minutes et 20 heures, et le BMI d'un individu doit se situer entre 10 et 65. \\

Nous sélectionnons 3 jeux de données, pour lesquels nous utlisons 10 mois consécutifs d'historique pour prédire les 2 prochains mois (voir l'annexe \ref{description_data} pour les statistiques descriptives des datasets):

\begin{itemize}
    \item Univarié : 
    \begin{itemize}
        \item un dataset de \textit{BMI} de 60 individus 
        \item un dataset de \textit{BMI} de 14 416 individus
    \end{itemize}
    \item Multivarié : un dataset de \textit{BMI} et de \textit{Sleep} de 14 416 individus
\end{itemize}
\ \\
Toutes les données sont standardisées\footnote{La standardisation ajuste la distribution d'une série temporelle en utilisant la moyenne et l'écart type du dataset d'entraînement, appliquant ensuite ces paramètres au dataset de test : $x_{standardise} = (x-moyenne)/ecart-type$.}, ce qui permet de favoriser la comparaison des deux variables qui n'ont pas la même échelle de mesure. Nous utilisons les mêmes individus dans tous les jeux de données de test, ce qui donne une taille constante de 18 individus pour notre ensemble de test, représentant 30\% de notre plus petit dataset.  

\paragraph{Méthodologie experimentale} Pour évaluer ces deux modèles, nous nous intéressons particulièrement au temps d'entrainement et aux performances prédictives (RMSE et MAE). \\
Pour pouvoir interpréter les performances de nos modèles, nous évaluons aussi 3 prédicteurs naïfs :
\begin{enumerate}
    \item \textsc{Last Value} : les prédictions prennent la valeur de la dernière observation 
    \item \textsc{Mean} : les prédictions prennent la valeur de la moyenne des observations
    \item \textsc{Median} : les prédictions prennent la valeur de la médiane des observations
\end{enumerate}
\ \\
Nous n'avons pas pour objectif d'analyser la réactivité des modèles face à la sparsité des séries temporelles, ainsi les jeux de données que nous utilisons n'ont pas de données manquantes.\\

Les deux modèles sont évalués sur la machine suivante (pas d'utilisation de GPU) : AMD Ryzen 7 pro 5850u avec radeon graphics × 16, Ubuntu Linux 22.04. \\

\subsection{Analyse des séries univariées}

\paragraph{Petit dataset}

\begin{table}[ht]
\begin{center}
\begin{tabular}{ |c|c|c|c|c|c|}
 \hline
 \multicolumn{2}{|c|}{\textsc{Last Value}} & \multicolumn{2}{|c|}{\textsc{Mean}} & \multicolumn{2}{|c|}{\textsc{Median}} \\
 \hline
 RMSE & MAE & RMSE & MAE & RMSE & MAE \\
 \hline
 $0.0608$ & $0.0557$ & $0.6439$ & $0.6425$ & $0.6252$ & $0.6229$ \\
 \hline
\end{tabular}
\caption{Résultats des modèles naïfs sur un dataset univarié de 60 individus.} \label{naive_univariate_small}
\ \\
\begin{tabular}{ |c||c|c|c|c|c|c|c|}
 \hline
 \multirow{2}{*}{k} & \multicolumn{3}{|c|}{\textsc{MagmaClust}} & \multicolumn{3}{|c|}{\textsc{DGM²}}\\
 \cline{2-7}
 & RMSE & MAE & Temps & RMSE & MAE & Temps \\
 \hline
 3  & $0.8061$ & $0.7344$ & 1 min 53 & $0.6693$ & $0.4643$ & 4 sec \\
 5  & $0.8767$ & $0.7890$ & 2 min 32 & $0.6967$ & $0.4721$ & 4 sec \\
 7  & $0.8016$ & $0.7275$ & 2 min 27 & $\bold{0.6400}$ & $\bold{0.4580}$ & 4 sec \\
 10 & $0.8019$ & $0.7274$ & 3 min 37 & $0.7219$ & $0.4995$ & 4 sec \\
 \hline
\end{tabular} 
\caption{Résultats des deux modèles comparés sur un dataset univarié de 60 individus.} \label{tab_univariate_small}
\end{center}
\end{table}

Malgré leur bonne performance en prédiction, les modèles naïfs ne nous aident pas à regrouper les données en clusters, ce qui est notre but. Nous les utilisons pour avoir un point de comparaison pour évaluer la performance de nos modèles. Par ailleurs, puisque \textit{BMI} est une variable plutôt stable, nous observons dans le Tableau \ref{naive_univariate_small} que l'estimateur naïf \textsc{Last Value} a de très bonnes performances. \\

Le Tableau \ref{tab_univariate_small} montre que sur un petit dataset univarié, \textsc{DGM²} a de meilleurs performances que \textsc{MagmaClust} (en moyenne 0.14 d'écart RMSE et 0.27 MAE), bien que ces résultats soient inférieurs à ceux des estimateurs naïfs. Nous remarquons par ailleurs que \textsc{DGM²} est bien plus rapide, contrairement à l’attente commune que
les approches de deep learning sont souvent plus gourmandes en ressources. \textsc{MagmaClust} utilise quant à lui un algorithme Variationel Espérance-Maximisation (VEM) de complexité $ \mathcal{O} (M \times N^3 + K \times N^3)$, avec $M$ le nombre d'individus, $N$ le nombre de points de temps et $K$ le nombre de clusters.

\paragraph{Grand dataset}
\begin{table}[ht]
\begin{center}
\begin{tabular}{ |c|c|c|c|c|c|}
 \hline
 \multicolumn{2}{|c|}{\textsc{Last Value}} & \multicolumn{2}{|c|}{\textsc{Mean}} & \multicolumn{2}{|c|}{\textsc{Median}} \\
 \hline
 RMSE & MAE & RMSE & MAE & RMSE & MAE \\
 \hline
 $0.0675$ & $0.0611$ & $0.7025$ & $0.7006$ & $0.7039$ & $0.7023$ \\
 \hline
\end{tabular} 
\caption{Résultats des modèles naïfs sur un dataset univarié de 14 416 individus.} \label{naive_univariate_big}
\ \\
\begin{tabular}{ |c||c|c|c|c|c|c|c|}
 \hline
 \multirow{2}{*}{k} & \multicolumn{3}{|c|}{\textsc{MagmaClust}} & \multicolumn{3}{|c|}{\textsc{DGM²}}\\
 \cline{2-7}
 & RMSE & MAE & Temps & RMSE & MAE & Temps \\
 \hline
 3  & $0.8100$ & $0.7589$ & 14h58 & $0.4219$ & $0.3477$ & 9 min 45  \\
 5  & $0.6690$ & $0.6008$ & 13h08 & $0.2995$ & $0.2295$ & 10 min 37 \\
 7  & $0.6732$ & $0.6041$ & 19h22 & $0.2743$ & $0.2126$ & 11 min 05 \\
 10 & $0.6734$ & $0.6201$ & 27h18 & $\bold{0.2068}$ & $\bold{0.1590}$ & 11 min 32 \\
 \hline
\end{tabular} 
\caption{Résultats des deux modèles comparés sur un dataset univarié de 14 416 individus.} \label{tab_univariate_big}
\end{center}
\end{table}

Nous constatons dans le Tableau \ref{tab_univariate_big} que \textsc{MagmaClust} gagne légèrement en performance, et la durée nécessaire pour l'entrainement sur un dataset de cette taille est notable. Une forte amélioration des performances du modèle \textsc{DGM²} est visible avec notre dataset d'entraînement plus grand 
(gain de 0.4332 de RMSE et 0.299 de MAE en prenant les modèles avec les meilleures performances pour chaque dataset). Cette progression souligne l'efficacité de ce modèle à saisir la structure latente des clusters lorsqu'il a appris sur suffisament de données, ce qui se traduit par des prédictions nettement plus précises par rapport à celles issues des modèles prédictifs naïfs \textsc{Mean} et \textsc{Median} visibles dans le Tableau \ref{naive_univariate_big}.

\subsection{Analyse étendue au cadre multivarié}

\begin{table}
\hspace*{-0.6cm}
\begin{centering}
\begin{tabular}{ |c|c|c|c|c|c|c|c|c|c|}
 \hline
 \multicolumn{4}{|c|}{\textsc{DGM²} multivarié} & \multicolumn{5}{|c|}{\textsc{DGM²} univariés combinés} & \multirow{2}{*}{ARI} \\
 \cline{1-9}
 $k$ & RMSE$_{m}$ & MAE$_{m}$ & Temps & $k_{BMI}$ & $k_{Sleep}$ & RMSE & MAE & Temps & \\
 \hline
 15 & $0.593$ & $0.443$ & 11 min 01 & $5$ & $3$ & $0.592$ & $0.445$ & 19 min 28 & $0.103$ \\
 25 & $0.591$ & $0.417$ & 12 min 40 & $5$ & $5$ & $0.564$ & $0.402$ & 19 min 46 & 0.261\\
 35 & $0.550$ & $0.400$ & 14 min 17 & $7$ & $5$ & $0.552$ & $0.394$ & 20 min 14 & 0.174\\
 49 & $0.563$ & $0.398$ & 17 min 10 & $7$ & $7$ & $0.541$ & $0.401$ & 20 min 31 & 0.236\\
 \hline
\end{tabular} 
\caption{Résultats du modèle \textsc{DGM²} sur un dataset multivarié et de la somme de l'agglomération de deux modèles \textsc{DGM²} univariés, sur un dataset de 14 416 individus.} \label{tab_multivariate_big}
\end{centering}
\end{table}

L'emploi d'un modèle de clustering multivarié, par rapport à des modèles univariés combinés, permet de capturer les interactions entre variables, essentielles à la compréhension de phénomènes complexes.  \textsc{MagmaClust} ne gérant pas la prédiction multivariée, nous testons donc seulement les performances de \textsc{DGM²} sur notre dataset multivarié.\\

Nous souhaitons nous assurer que l'introduction de la dimension multivariée ne diminue pas la qualité des prédictions par rapport à l'utilisation de deux modèles univariés de complexité équivalente. Pour cela, on compare les résultats d'un modèle multivarié à $k$ clusters deux modèles univariés chacun spécialisé dans une variable telle que : $k = k_{BMI} \times k_{Sleep}$. Il convient de souligner que les clusters formés par le modèle multivarié visent à saisir la structure latente partagée simultanément par les variables \textit{BMI} et \textit{Sleep}, et ne sont donc pas directement comparables à ceux obtenus dans les modèles univariés, comme le montre l'ARI faible entre les deux approches dans le Tableau \ref{tab_multivariate_big}. C'est aussi dû à cet obstacle dans la comparaison qu'est représenté dans la colonne "\textsc{DGM²} multivarié" la moyenne des RMSE et des MAE pour chaque variable dans le modèle multivarié \footnote{Ce qui n'est pas équivalent au RMSE du modèle multivarié, mais équivalent à son MAE.}. Le détail des résultats de ce modèle est disponible dans le Tableau \ref{tab_multivariate_details}. La colonne "\textsc{DGM²} univariés combinés", représente la moyenne des RMSE et MAE des prédictions des deux modèles univariés, dont les résultats détaillés sont disponibles dans le Tableau \ref{tab_multi_univarite}. Nous constatons que les performances des deux approches sont équivalentes \footnote{Des analyses supplémentaires, non exposés dans cet article, suggèrent que la différence des performances de prédiction entre le modèle multivarié et les modèles univariés combinés n'est pas significative.}. De plus, l'utilisation d'un unique modèle multivarié se révèle être plus efficiente en termes de temps de calcul que l'exécution séparée de deux modèles univariés.

\section{Conclusion}

Le clustering dynamique pour suivre l'évolution de l'état d'un patient est pertinent dans notre cadre d'application, permettant de saisir des variations subtiles et des changements progressifs dans les données temporelles, offrant une perspective plus détaillée de la santé du patient. Le modèle \textsc{DGM²} se distingue par sa capacité à identifier ces clusters dynamiques, par ses performances prédictives et sa rapidité d'entraînement par rapport à \textsc{MagmaClust}. \\

Cependant, choisir le nombre approprié de clusters est un défi, car un nombre plus élevé rend l'interprétation plus complexe, tandis qu'un nombre insuffisant peut masquer des informations intéressantes, simplifiant excessivement la dynamique des données. Il est donc crucial de trouver un équilibre entre la décomposition détaillée des données et la facilité d'interprétation clinique. Trop de clusters peuvent nuire à l'analyse des tendances et à la prise de décision, alors qu'un nombre insuffisant peut ignorer des aspects importants de l'évolution du patient. \\

Dans la continuité des résultats prometteurs de \textsc{DGM²}, qui présente des avantages significatifs pour le clustering prédictif en suivi patient, nos perspectives de recherche s’orientent vers une exploration plus approfondie du clustering dynamique. Nous sommes particulièrement encouragés par les performances computationnelles observées, qui valident l’emploi de méthodes de deep learning pour cette tâche. De plus, nous anticipons une amélioration des capacités prédictives du modèle multivarié en présence de corrélations entre les dimensions.

\vspace{-0.7cm}
\bibliographystyle{rnti}
\bibliography{biblio}

\appendix
\newpage
\section{Annexe}

\subsection{Description des données} \label{description_data}

\begin{figure}[ht]
\begin{center}
 \includegraphics[width=5cm]{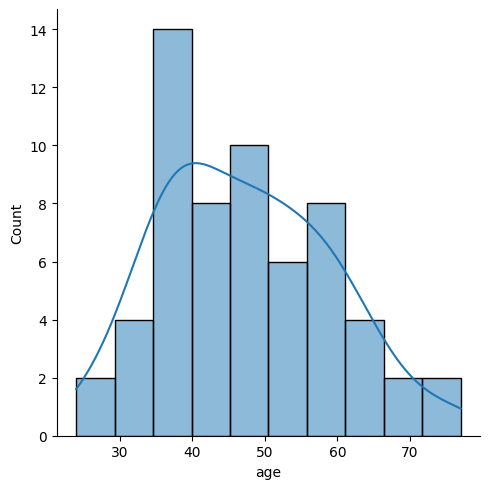}
 \caption{Histogramme des âges des individus du petit dataset.} \label{hist_age_petit}
\end{center}
\end{figure}

\begin{figure}[ht]
\begin{center}
 \includegraphics[width=9cm]{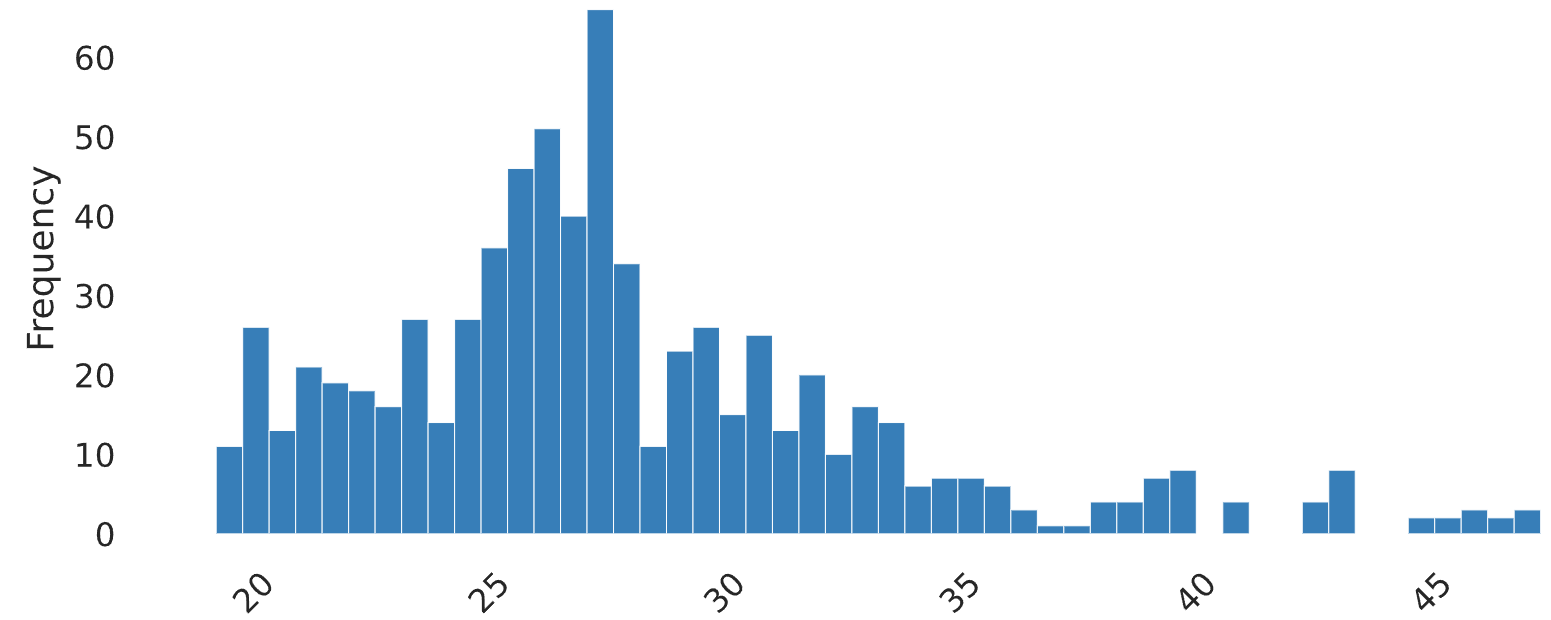}
 \caption{Histogramme des valeurs de \textit{BMI} des individus du petit dataset.} \label{hist_bmi_petit}
\end{center}
\end{figure}

\paragraph{Petit dataset} Le dataset de 60 individus. Il y a 85\% d'hommes (51) et 15\% de femmes (9). La distribution de l'âge est exposé dans l'histogramme sur la Figure \ref{hist_age_petit}. La distribution des valeurs de la variable \textit{BMI} des individus du petit dataset est visible sur la Figure \ref{hist_bmi_petit}.

\begin{figure}[ht]
\begin{center}
 \includegraphics[width=5cm]{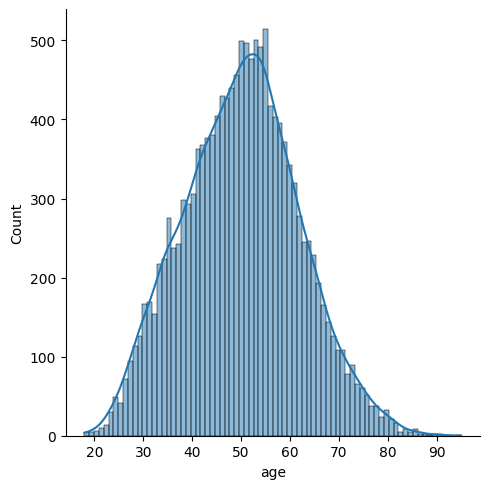}
 \caption{Histogramme des âges des individus du grand dataset.} \label{hist_age_grand}
 \ \\
 \includegraphics[width=8cm]{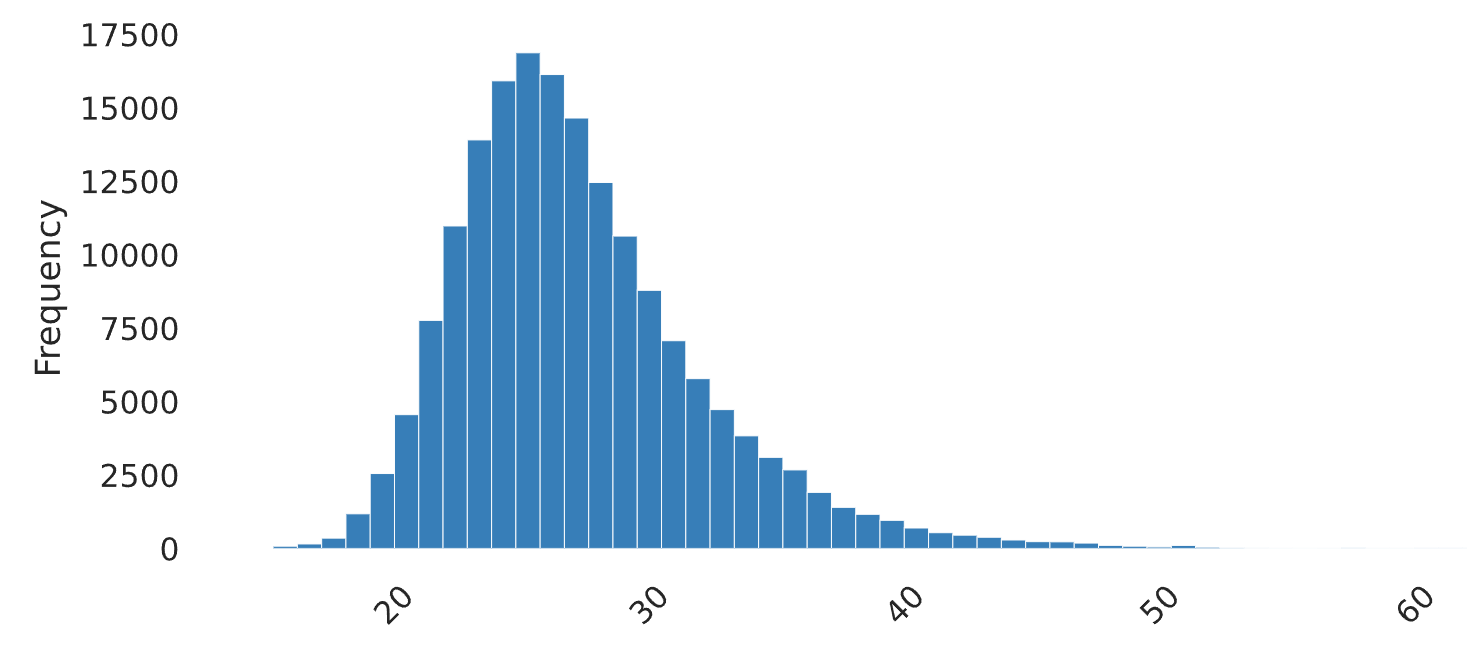}
 \caption{Histogramme des valeurs de \textit{BMI} des individus du grand dataset.} \label{hist_bmi_grand}
 \ \\
 \includegraphics[width=7.5cm]{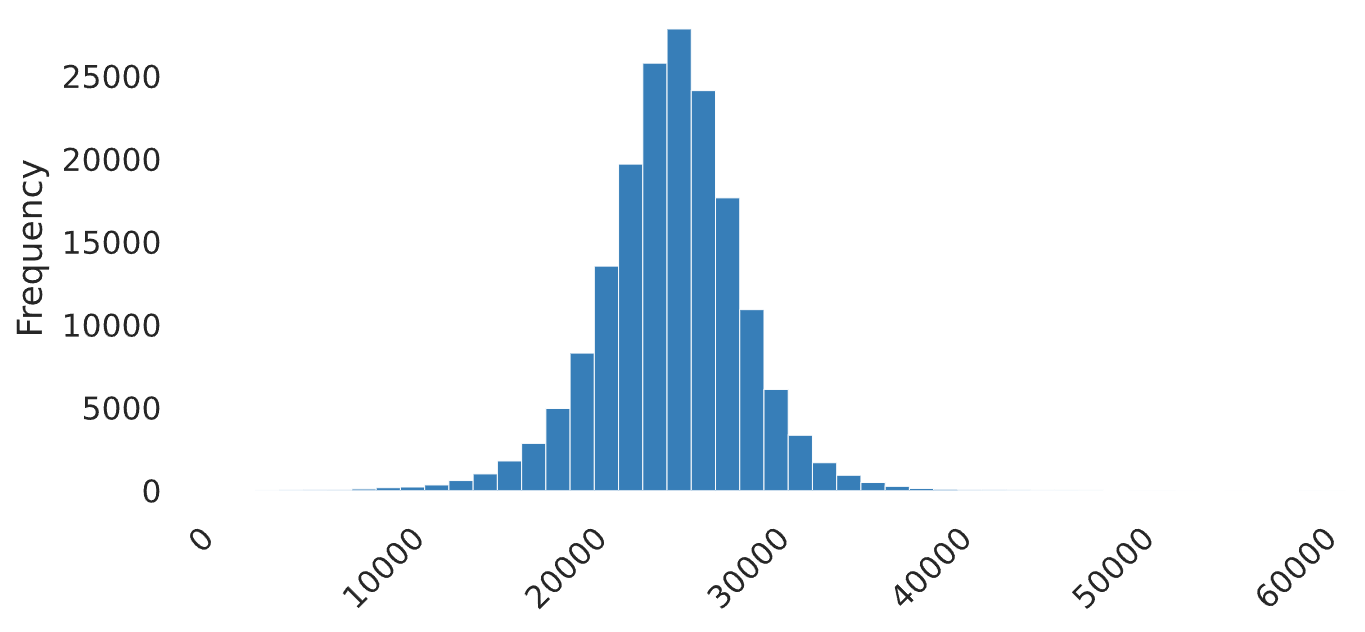}
 \caption{Histogramme des valeurs de \textit{Sleep} des individus du grand dataset.} \label{hist_sleep_grand}
\end{center}
\end{figure}

\newpage
\paragraph{Grand dataset} Le dataset de  14 416 individus. Il y a 84,86\% d'hommes (12 233) et 15,14\% de femmes (2 182). La distribution des âges est exposée sur la Figure \ref{hist_age_grand}. La distribution de la variable \textit{BMI} des individus du grand dataset est visible sur la Figure \ref{hist_bmi_grand} et celle de la variable \textit{Sleep} sur la Figure \ref{hist_sleep_grand}.

\subsection{Description du mélange de distribution gaussiennes dynamique} \label{dynamique}

\begin{figure}[t]
\begin{center}
 \includegraphics[width=5cm]{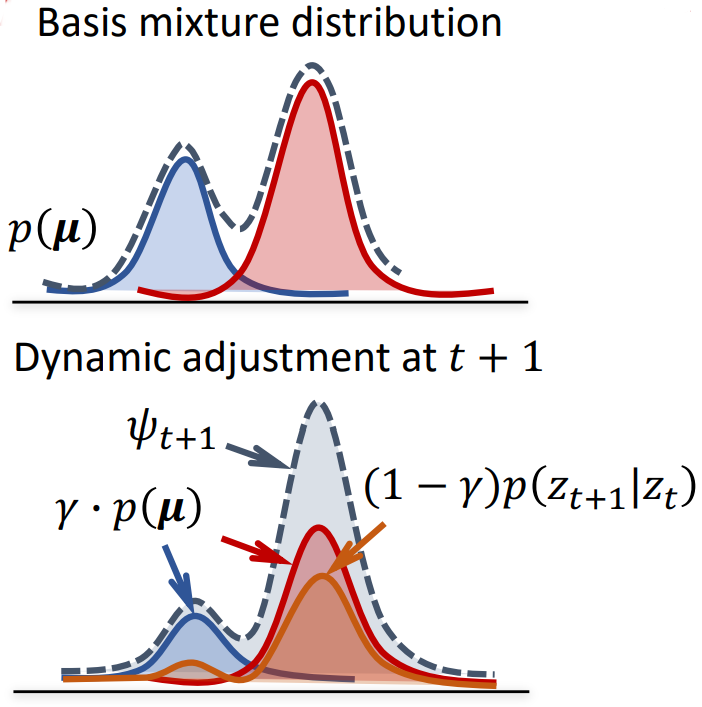}
 \caption{Ajustement dynamique du mélange gaussian selon l'équation (\ref{eq:dynamic}), avec 2 composantes.} \label{fig:dynamic}
\end{center}
\end{figure}

Soit $\mu_k$, avec $k=\{1,...,K\}$, la moyenne de la $k$-ème composante du mélange gaussien (statique), et soit $p(\mu_k)$ sa probabilité correspondante. Soit $p(\mu) = [p(\mu_1), ..., p(\mu_K)] \in \mathbb{R}^K$. On utilise (dans l'étape 3) le mélange de distribution dynamique suivant : 
\begin{align}\label{eq:dynamic}
    \psi_{i,t+1} = (1 - \gamma) p (z_{i,t+1}|z_{i,1:t})+ \gamma p(\mu)
\end{align}
avec $\gamma \in [0,1]$ un hyperparamètre qui contrôle le degré relatif de changement par rapport à la distribution de mélange statique. La Figure \ref{fig:dynamic} illustre ce processus.

\subsection{Détails des résultats de DGM² multivarié}

\begin{table}[h]
\begin{center}
\begin{tabular}{ |c||c|c|c|c|}
 \hline
 & RMSE BMI& MAE BMI & RMSE Sleep & MAE Sleep \\
 \hline
 \textsc{Last Value} & $0.0675$ & $0.0610$ & $0.7983$ & $0.7141$ \\
 \textsc{Mean}       & $0.7024$ & $0.7006$ & $0.9872$ & $0.9259$ \\
 \textsc{Median}     & $0.7038$ & $0.7022$ & $0.9794$ & $0.9179$ \\
 \hline
\end{tabular}
\caption{Résultats des modèles naïfs sur un dataset multivarié de 14 416 individus.} \label{naif_multivariate_big}
\end{center}
\end{table}

On constate dans le Tableau \ref{naif_multivariate_big} que sur une variable moins stable que le \textit{BMI} comme la variable \textit{Sleep}, le prédicteur naïf \textsc{Last Value} est moins performant. \\

\begin{table}[ht]
\begin{center}
\begin{tabular}{ |c||c|c|c|c|c|c|c|c|c|}
\hline
k  & RMSE$_{multi}$ & MAE$_{multi}$ & RMSE BMI & MAE BMI & RMSE Sleep & MAE Sleep & Temps \\
\hline
15 & $0.6844$ & $0.4438$ & $0.2537$ & $0.2014$ & $0.9340$ & $0.6863$ & 11 min 01 \\
21 & $0.6979$ & $0.4406$ & $0.2277$ & $0.1810$ & $0.9604$ & $0.7002$ & 12 min 39 \\
25 & $0.6746$ & $0.4173$ & $0.2683$ & $0.1958$ & $0.9156$ & $0.6387$ & 12 min 40 \\
35 & $0.6497$ & $0.4004$ & $0.2047$ & $0.1464$ & $0.8957$ & $0.6544$ & 14 min 17 \\
49 & $0.6471$ & $0.3986$ & $0.2450$ & $0.1626$ & $0.8816$ & $0.6347$ & 17 min 10 \\
\hline
\end{tabular}
\caption{Résultats du modèle \textsc{DGM²} sur un dataset multivarié de 14 416 individus.} \label{tab_multivariate_details}
\end{center}
\end{table}

Les "RMSE$_{moy}$" et "MAE$_{moy}$" du Tableau \ref{tab_multivariate_big} proviennent de la moyenne des colonnes "RMSE BMI" et "MAE BMI" du Tableau \ref{tab_multivariate_details}.

\begin{table}[ht]
\begin{center}
\begin{tabular}{ |c||c|c|c|c|c|c|}
\hline
\multirow{2}{*}{k} & \multicolumn{3}{|c|}{\textit{Sleep}} & \multicolumn{3}{|c|}{\textit{BMI}}\\
\cline{2-7}
 & RMSE & MAE & Temps & RMSE & MAE & Temps \\
\hline
3 & $0.8858$ & $0.6606$ & 8 min 51 & $0.4219$ & $0.3477$ & 9 min 45  \\
5 & $0.8304$ & $0.5754$ & 9 min 09 & $0.2995$ & $0.2295$ & 10 min 37 \\
7 & $0.8084$ & $0.5910$ & 9 min 26 & $0.2743$ & $0.2126$ & 11 min 05 \\
\hline
\end{tabular}
\caption{Résultats du modèle \textsc{DGM²} sur un dataset univarié \textit{Sleep} et un dataset univarié \textit{BMI} de 14 416 individus.} \label{tab_multi_univarite}
\end{center}
\end{table}

Les "RMSE" et "MAE" de la colonne "DGM² univariés combinés" du Tableau \ref{tab_multivariate_big} proviennent de la moyenne des colonnes RMSE et MAE des deux variables du Tableau \ref{tab_multi_univarite}.

\Fr

\end{document}